\documentclass{article}

\usepackage{arxiv}

\usepackage[utf8]{inputenc} % allow utf-8 input
\usepackage[T1]{fontenc}    % use 8-bit T1 fonts
\usepackage{hyperref}       % hyperlinks
\usepackage{url}            % simple URL typesetting
\usepackage{booktabs}       % professional-quality tables
\usepackage{amsfonts}       % blackboard math symbols
\usepackage{nicefrac}       % compact symbols for 1/2, etc.
\usepackage{microtype}      % microtypography
\usepackage{lipsum}
\usepackage{graphicx}
\usepackage[toc,page]{appendix}
\usepackage{xcolor}

\def\eg{\emph{e.g}}

\def\etal{\emph{et al}}

\title{Attribute-Guided \\ Image Generation from Layout}

\author{
  Ke MA  \\
  Department of Computer Science\\
  University of British Columbia\\
  Vancouver, Canada \\
  \texttt{mark1123@cs.ubc.ca} \\
   \And
 Bo Zhao \\
  Bank of Montreal\\
  Toronto, Canada \\
  \texttt{zhaobo.cs@gmail.com} \\
   \AND
    Leonid Sigal  \\
  Department of Computer Science\\
  University of British Columbia\\
  Vancouver, Canada \\
  \texttt{lsigal@cs.ubc.ca} \\
}

\begin{document}
\maketitle

\begin{abstract}
    Recent approaches have achieved great successes in image generation from structured inputs, e.g., semantic segmentation, scene graph or layout. Although 
    these methods allow specification of objects and their locations at image-level, they lack the fidelity and semantic control to specify visual appearance of these objects at an  instance-level.
    To address this limitation, 
    we propose a new image generation method that enables instance-level attribute control. 
    Specifically, the input to our attribute-guided generative model is a tuple that contains: (1) object bounding boxes, (2) object categories and (3) an (optional) set of attributes for each object. The output is a generated image where the requested objects are in the desired locations and have prescribed attributes.
    Several losses work collaboratively to encourage accurate, consistent and diverse image generation.
    Experiments on Visual Genome \cite{krishna2017visual} dataset demonstrate our model's capacity to control object-level attributes in generated images, and validate plausibility of 
    disentangled object-attribute representation in the image generation from layout task. Also, the generated images from our model have higher resolution, object classification accuracy and consistency, as compared to the previous state-of-the-art. 
\end{abstract}

%-------------------------------------------------------------------------
\section{Introduction}
\label{sec:intro}
Controlled image generation methods have achieved great successes in recent years, driven by the advances in conditional Generative Adversarial Networks (GANs)~\cite{goodfellow2014generative, kim2017learning, miyato2018cgans, odena2017conditional, reed2016generative, zhang2018self, zhang2017stackgan, zhu2017unpaired, zhu2017toward} and disentangled representations~\cite{kazemi2019style, zhu2018visual}. The goal of these methods is to generate high-fidelity images from various user specified guidelines (conditions), such as textual descriptions \cite{hong2018inferring, mansimov2015generating, tan2018text2scene, xu2018attngan,zhang2017stackgan}, attributes \cite{dong2017semantic, karacan2016learning, li2019attribute, lu2018attribute, nam2018text, zhou2019text},  scene graphs \cite{ashual2019specifying, johnson2018image, li2019pastegan}, layout \cite{sun2019image, zhao2019image} and semantic segmentation \cite{chen2017photographic, huang2018multimodal, isola2017image,karacan2016learning, liu2017unsupervised, park2019semantic,wang2018high, zhu2017unpaired, zhu2017toward}.  
The high-level nature of most of these specifications is desirable from ease of use and control point of views, but severely impoverished in terms of pixel-level spatial and appearance information, leading to a challenging image generation problem. 

In this paper, we specifically focus on image generation from layout, where the input is a course spatial layout of the scene (\eg, bounding boxes and corresponding object class labels) and the output is an image consistent with the specified layout. Compared to text-to-image~\cite{reed2016generative, zhang2017stackgan} and scene-graph-to-image~\cite{ashual2019specifying, johnson2018image, li2019pastegan} generation paradigms, layout-to-image provides an easy, spatially aware and interactive abstraction for specifying desired content. This makes this paradigm compelling and effective for users across the spectrum of potential artistic skill sets; from children and amateurs to designers. 
Image generation from a layout is a relatively new problem, with earlier methods using layout only as an intermediate representation \cite{hong2018inferring, johnson2018image}, but not a core abstraction or specification exposed to the user.

\begin{figure}[t]
    \begin{center}
        \includegraphics[trim=0in 0.1in 0in 0in, clip, width=0.62\linewidth]{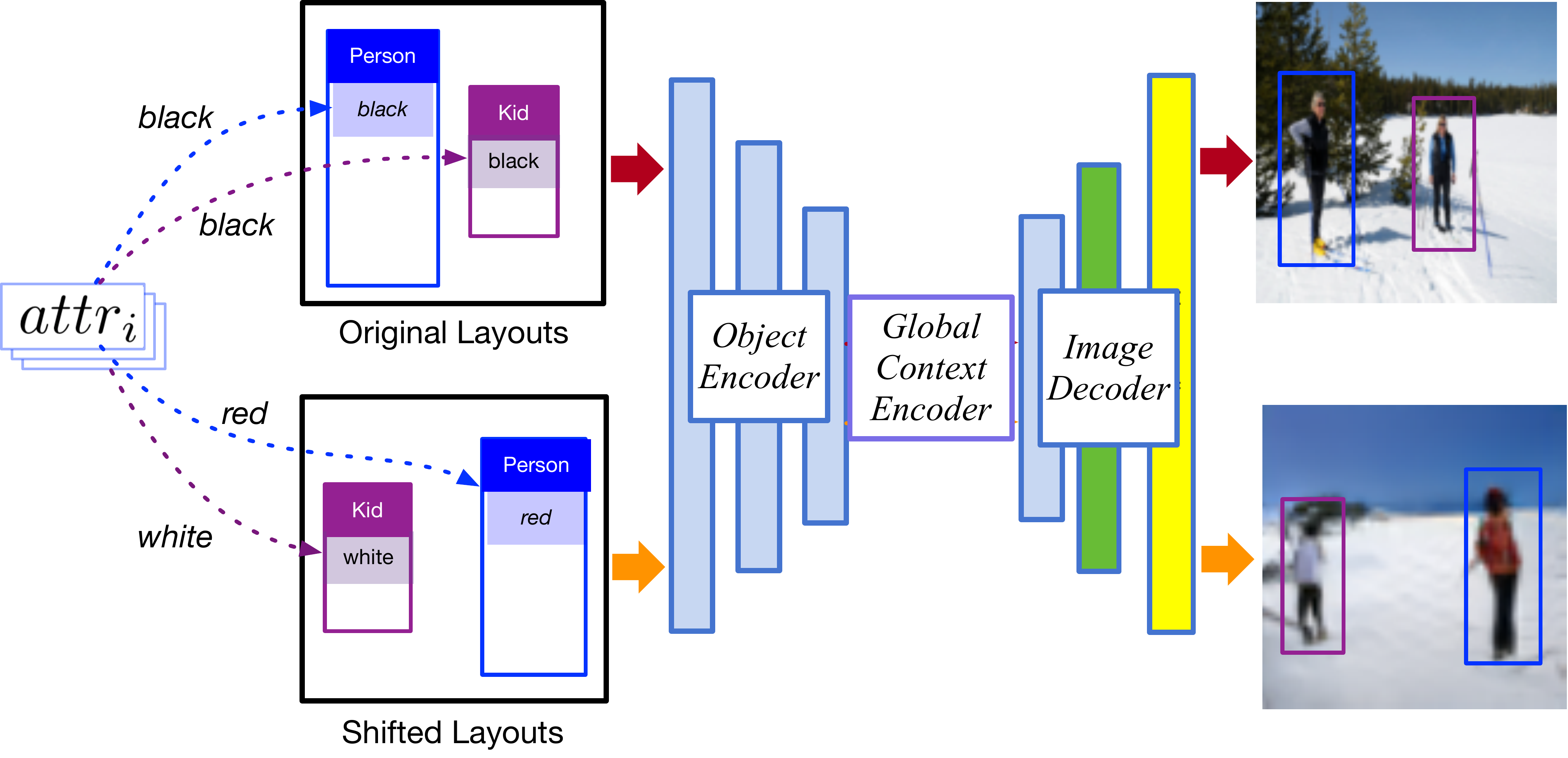}
    \end{center}
    \vspace{-0.2in}
    \caption{{\bf Attribute-guided Image Generation from Layout.} Unlike prior layout-based image generation architectures, our model allows for instance-level granular semantic attribute control over individual objects  (\eg, specifying that a {\tt person} should be wearing something {\tt black} (top) or {\tt red} (bottom)); it
    also ensures appearance consistency when bounding boxes in layout undergo translation.}
    \label{fig:itro}
    \vspace{-0.15in}
\end{figure}

Layout2Im~\cite{zhao2019image} was the first model proposed for image generation from layout, followed by more recent LostGAN~\cite{sun2019image}, which improved on the performance in terms of overall image %  generation 
quality. However, all current image generation from layout frameworks \cite{sun2019image, zhao2019image} are limited in a couple of fundamental ways. 
First, they {\em lack ability to semantically control individual object instances}. While both Layout2Im and LostGAN model distributions over appearances of objects through appearance \cite{zhao2019image} or style \cite{sun2019image} latent codes, neither is able to control these variations semantically.
% in any meaningful or semantic manner. 
One can imagine using encoded sample patches depicting desired objects as an implicit {\em control} mechanism ({\em i.e.}, generate an instance of a {\tt tree} or {\tt sky} that resembles an example in a given image patch),
%an instance in this image patch, or make {\tt sky} resemble the rectangular sky patch of this other image), 
however, this is in the very least awkward and time consuming from the user perspective.
Second, they generally {\em lack consistency} -- are not spatially equivariant. Intuitively, shifting a location (bounding box) of an object in the layout specification, while keeping appearance/style latent code fixed, should result in the object simply shifting by the relatively same amount in the output image (property known as equivariance).
% more generally known in mathematics as equivariance). 
However, current models fail to achieve this % , from the use point of view, 
intuitive consistency. 
Finally, they are {\em limited to low-resolution} output images, typically of size 64$\times$64. 

In this paper, we address these challenges by proposing a new framework for attribute-guided image generation from layout, building on, and substantially extending, the backbone of \cite{zhao2019image}. In particular, we show that a series of simple and intuitive architectural changes: incorporating (optional) attribute information, adopting a global context encoder, and adding additional image generation path where object locations are shifted -- leads to the instance-level fine-grained control over the generation of objects, while increasing the image quality and resolution. We call this model attribute-guided layout2im (see Figure~\ref{fig:itro}). 

\vspace{0.07in}
\noindent
{\bf Contributions:} Our contributions are three fold: (1) our attribute-guided layout2im architecture allows (but does not require) direct instance-level granular semantic attribute control over individual objects; (2) is directly optimized to be consistent, or equivariant, with respect to spatial shifts of object bounding boxes in the input layout; and (3) allows for higher-resolution output images of size up to 128$\times$128 by utilizing global context encoder and progressive upsampling. Both qualitatively and quantitatively we show state-of-the-art generative  performance on Visual Genome \cite{krishna2017visual} benchmark dataset, while benefiting from the desirable control properties, unavailable in other models. The code and pretrained models will be made available \footnote{\tiny \url{https://github.com/UBC-Computer-Vision-Group/attribute-guided-image-generation-from-layout}}.

%------------------------------------------------------------------------
\section{Related Work}
\noindent
{\bf Image Generation from Scene Graph:}
Scene graph is a convenient directed graph structure designed to represent a scene,  where nodes correspond to objects and edges to the relationships between objects. Recently, scene graphs have been used in many image generation tasks due to their flexibility in explicitly encoding object-object relationships and interactions~\cite{ashual2019specifying, johnson2018image, li2019pastegan}. The typical generation process involves two phases: (1) a graph convolutional neural network~\cite{henaff2015deep} is applied to the scene graph to predict the scene layout ({\em i.e.}, bounding boxes and segmentation masks); and (2) the layout is then decoded into an image. Different from methods that generate image as a whole, Li \etal~\cite{li2019pastegan} propose a semi-parametric approach and crop refining network to reconcile the isolated crops into an integrated image. 
% We note that 
Unlike in our approach, the scene layout in these models is used as an  intermediate semantic representation to bridge abstract scene graph input and an image output. 
%   As compared to scene graph, bounding boxes capture object-object relationships in a easier and more interactive manner. Hence, we choose bounding boxes as our input. 
% through networks such as auto-encoder and refinement networks. 
% Visual Genome ~\cite{krishna2017visual} and COCO-stuff datasets ~\cite{lin2014microsoft} are mostly used for this task. 

\vspace{0.06in}
\noindent
{\bf Image Generation from Layout:}
Image layout, comprising bounding box locations, often serves as an intermediate step for image generation (see above). Zhao \etal~\cite{zhao2019image} proposed image generation from layout as a task in its own right, where the image is generated from bounding boxes and corresponding object categories. To combine multiple objects, ~\cite{zhao2019image} sequentially fuse object feature maps using a convolutional LSTM (cLSTM) network; the resulting fused feature map is then decoded to an image. Turkoglu \etal~\cite{turkoglu2019layer}, on the other hand, divide the generation into multiple stages where objects are added to the canvas one by one. To better bridge the gap between layouts and images, Li \etal~\cite{li2019object} uses a shape generator to outline the object shape and provide the model fine-grained information from text using object-driven attention. Similarly, ~\cite{sun2019image} learns object-level fine-grained mask maps that outline the object shape. In addition,  ~\cite{karras2019style, sun2019image} show that incorporating layout information into normalization layer yields better results: adopting instance normalization technique ~\cite{karras2019style} in generator realize multi-object style control ~\cite{sun2019image}, whereas spatially-adaptive normalization ~\cite{park2019semantic} based on segmentation mask modulates activation in upsampling layers. Taking the inspiration from ~\cite{ karras2019style, park2019semantic,  sun2019image}, we apply spatially-adaptive normalization in our generator to compute layout-aware affine transformations in normalization layers.

\begin{figure*}[!ht]
  \centering
  \includegraphics[width=\linewidth]{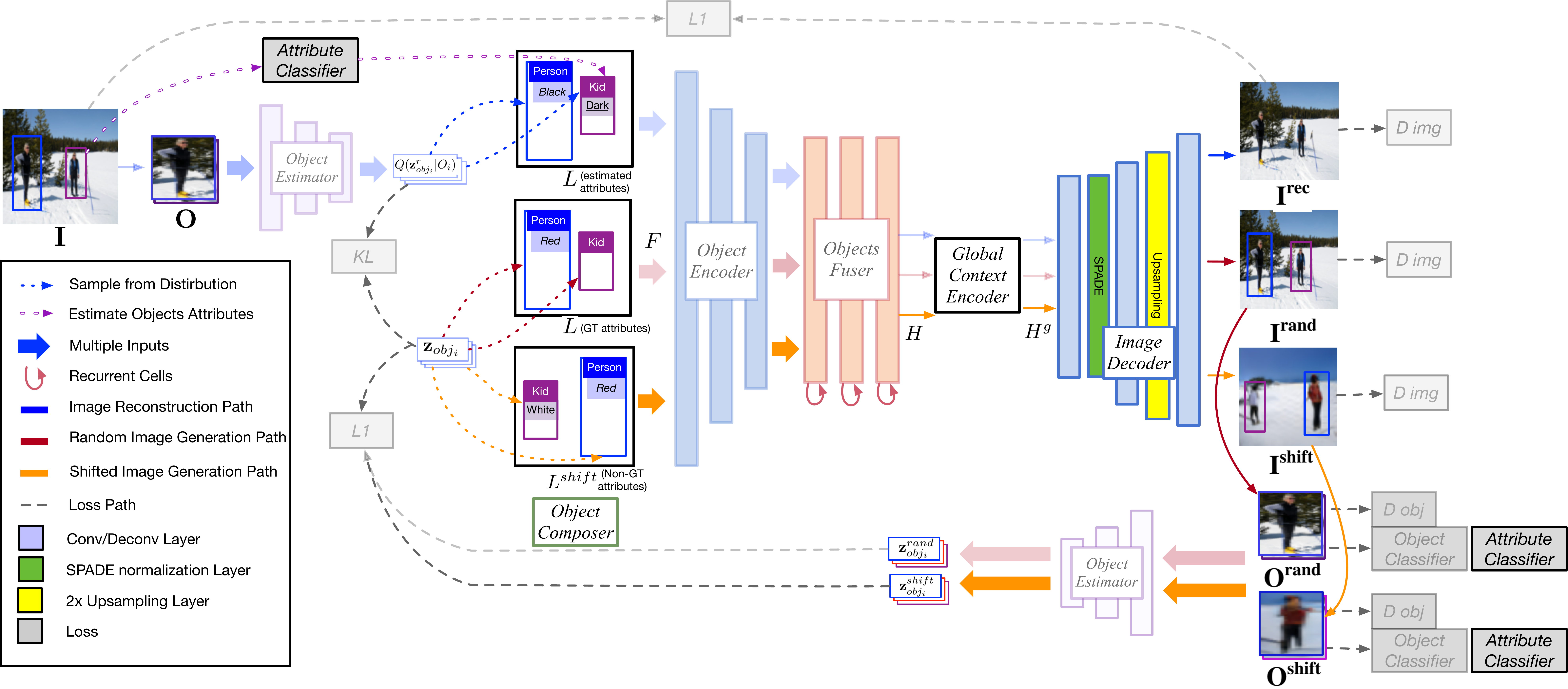}
  \caption{{\bf Overview of our Attribute-Guided Layout2Im training pipeline}. Our architecture has three generation paths: Image reconstruction path (top/blue), Image generation path (middle/red) and Layout reconfiguration path (bottom/orange). Attribute classifier is used in reconstruction path to estimate attributes of objects that do not have any attribute annotations.  Non-GT (sampled) attributes are used in image generation path and layout reconfiguration path to disentangle attribute information from class ($\mathbf{w}_i$) and appearance ($\mathbf{z}_{obj_i}$).}
  % $w$ and $z_{obj}$. 
  % See text for details. }
  \vspace{-0.1in}
  \label{fig:pipeline}
%   \vspace{-0.15in}
\end{figure*}

\vspace{0.06in}
\noindent
{\bf Semantic Image Synthesis:}
Semantic image synthesis is an image-to-image translation task. While most methods use conditional adversarial training \cite{goodfellow2014generative, mirza2014conditional}, such as pix2pix \cite{isola2017image}, pix2pixHD \cite{wang2018high}, cVAE-GAN and cLR-GAN \cite{zhu2017toward}, others such as Cascaded Refinement Networks \cite{chen2017photographic} also yields plausible results. To preserve semantic information, normalization techniques like SPADE  \cite{park2019semantic} have recently been deployed to modulate the activations in normalization layers through a spatially adaptive and learned transformation. Semantic image synthesis can also serve as an intermediate step for image modeling \cite{wang2016generative}. In addition, some image-to-image translation tasks are achieved in unsupervised manner \cite{huang2018multimodal, liu2017unsupervised, zhu2017unpaired}, but most existing semantic image synthesis models require paired data. 
%  Semantic masks, however, can be expensive to acquire, so our generator only relies on course layouts as the input. %  modality.

\vspace{0.06in}
\noindent
{\bf Attribute-guided Image Generation:}
In image generation,
various attempts have been made to specify the attributes of generated images and objects.
For example,~\cite{dong2017semantic, li2019attribute,lu2018attribute, nam2018text} aim to edit the attributes of the generated image using natural language descriptions. In~\cite{li2019attribute, zhou2019text} authors embed a visual attribute vector (\eg, blonde hair) for attribute-guided face generation. % However, their success is limited in bird, flower and human generation. 
Li \etal~\cite{li2019pastegan} also incorporates object-level appearance information in the input, but it relies on external memory to sample objects. Another line of the literature concentrates on editing images by providing reference images (\eg, \cite{ashual2019specifying}) to transfer style.
% ~\cite{karras2019style} enables scale-specific modifications of the synthesis, leading to excellent results in controlled face generation. 
% As an example, ~\cite{ashual2019specifying} extracts appearance attributes information from unrelated images, which are then applied to the generated image. 
Different from prior approaches, our model allows direct attribute control over  individual instances of objects in complex generative scenes.

\section{Approach}
Let us start by formally defining the problem. 
% In this section, we first define the problem and then introduce our proposed model.
%
% \subsection{Problem Formulation}
Let $\Lambda$ be an image canvas (\eg, of size 128$\times$128) and let $L$ = $\{ (\ell_i, \mathbf{attr}_i, \mathbf{bbox}_i)^m_{i=1} \}$ be a layout consisting of $m$ labeled object instances with class labels $ \ell_i \in \mathcal{C}$, attributes $\mathbf{attr}_i = \{a_{ij}\}_{j=1}^{n_i}$, 
% where $\mathcal{E}_{ij} \in \mathcal{A}$ 
and bounding boxes defined by top-left and  bottom-right coordinates on the canvas, $\mathbf{bbox}_i  \subseteq \Lambda$, where $|\mathcal{C}|$ is the total number of object categories and $a_{ij} \in \mathcal{A}$ is the $j$-th attribute of $i$-th object instance; $\mathcal{A}$ is the attribute set. Note that each object might have more than one attribute. Let $\mathbf{z}_{obj_i}$ be the latent code for object instance $(\ell_i, \mathbf{attr}_i, \mathbf{bbox}_i)$, modeling class- and attribute-unconstrained appearance.  We denote the set of all object instance latent codes in the layout $L$ by $Z_{obj} = \{\mathbf{z}_{obj_i}\}^m_{i=1}$.

Attribute-guided image generation from a layout can then be formulated as learning a generator function $G$ which maps given input $(L, Z_{obj})$ to the output image $I$ conforming to specifications:
\vspace{-0.05in}
\begin{equation} \label{eq:formulation}
  I = G(L, Z_{obj}; \Theta_G),
  \vspace{-0.05in}
  \end{equation}
where $\Theta_G$ parameterizes the generator function $G$ and corresponds to the set of  parameters that need to be learned. 

Different from previous layout to image generation methods is explicit, but optional ($\mathbf{att}_i$ can be $\emptyset$ or sampled from the prior), inclusion of the attributes. Further, we specifically aim to learn $G$ which is, at least to some extent, equivariant with respect to location of objects $\mathbf{bbox}_i$ in the layout.
% Specifically, we aim to learn $G$ such that it is 
%
Our attribute-guided layout2im formulation builds on and improves~\cite{zhao2019image}, as such it shares some of the basic architecture design principles outlined in Zhao \etal~\cite{zhao2019image}.  
%\subsection{Framework}
%Attribute-guided Layout2im is an improved model of layout2im, and thus it shares the basic structure of layout2im. 

\vspace{0.07in}
\noindent
{\bf Training:}
The overall training pipeline of the proposed approach is illustrated  in  Figure \ref{fig:pipeline}. Given the input image $I$ and its layout $L$ = $\{ (\ell_i, \mathbf{attr}_i, \mathbf{bbox}_i)^m_{i=1} \}$, our model first creates a word  embedding $\mathbf{w}_i$ for each object label $\ell_i$ and multi-hot attribute embedding $\mathbf{e}_i \in \{0,1\}^{|\mathcal{A}|}$ for object attribute(s)\footnote{Note exactly $n_i$ elements of $\mathbf{e}_i$ will be 1 and the rest are 0.} $\mathbf{attr}_i$, and form a  joint object-attribute embedding $M(\mathbf{w}_i \oplus \mathbf{e}_i)$ where $\oplus$ is the vector concatenation and $M$ is a MLP layer, composed of three fully connected layers that map the concatenated vector to a lower dimensional vector.  A set of object latent codes $Z_{obj} = \{\mathbf{z}_{obj_i}\}^m_{i=1}$ are sampled from the standard prior normal distribution $\mathcal{N}(0, 1)$, and a new $L^{shift}$ = $\{ (\ell_i, \mathbf{attr}_i, \mathbf{bbox}^{shift}_i)^m_{i=1} \}$ is constructed, where $\mathbf{bbox}^{shift}_i$ represents bounding boxes that are randomly shifted in the canvas $\Lambda$. The shifts  are horizontal to maintain consistency. Then, our model estimates another set of latent codes $Z^r_{obj} = \{\mathbf{z}^r_{obj_i}\}^m_{i=1}$, where $\mathbf{z}^r_{obj_i}$ is sampled from the posterior $Q(\mathbf{z}^r_{obj_i} | O_i)$ conditioned on CNN features of object $O_i$ cropped from input image $I$. % At the end 
We effectively end up with three datasets: 
\begin{itemize}
   \item[Set 1:] ($L$, $Z^{r}_{obj}$) for reconstruction of the original image. 
   Mapping this input through generator $G$
   should result in an % output 
   %  This input should result in an 
   image $I^{rec}$, which is a reconstruction of the training image $I$ serving as the source of the layout $L$; 
   \vspace{-0.06in}
   \item[Set 2:] ($L$, $Z_{obj}$) for generation of a new image sharing the original layout. The output of $G$ here should be image 
   $I^{rand}$ that shares the layout with above, but where appearance of each object instance is novel (sampled from the prior).
   %  is an image that shares the layout with above, but where appearance of each object instance is novel. We denote the resulting image $I^{rand}$ 
   \vspace{-0.06in}
  \item[Set 3:] ($L^{shift}$, $Z_{obj}$) which is used to generate an image from reconfigured layout ({\em i.e.}, reconfiguration path, see Suppl. Sec. 1.1 for details). The output should be a corresponding shifted image $I^{shift}$, which shares latent appearance codes with those from Set 2. 
\end{itemize}
 The same pipeline is applied to all three input sets simultaneously: multiple object feature maps $F_i$ are constructed based on the layout $L_i$ and $(\mathbf{z}_{obj_i} \oplus M(\mathbf{w}_i \oplus \mathbf{e}_i))$, and then fed into the object encoder and the objects fuser sequentially, generating a fused hidden feature map $H$ containing information from all specified objects. Lastly, we incorporate a global context embedding $\mathbf{g}$ onto $H$ to form a context-aware feature canvas $H^g$, and decode it back to an image with a combination of deconvolution, upsampling and SPADE normalization layers \cite{park2019semantic}. For generated object $O_i$ in $I^{rand}$ and $I^{shift}$, we make the object estimator regress the sampled latent codes $\mathbf{z}_{obj_i}$ based on $O_i$ to encourage $O_i$ to be consistent with the latent code $\mathbf{z}_{obj_i}$, and use an auxiliary object classifier and attribute classifier to ensure $O_i$ has the desired category and attributes. To train the model adversarially,  we also introduce a pair of discriminators, $D_{img}$ and $D_{obj}$, to classify results as being real/fake at image and object level. 

\vspace{0.07in}
\noindent
{\bf Inference:} At inference time, the model is able to synthesize a realistic image from a given (user specified) layout $L$ and object latent codes $Z_{obj}$ sampled from the prior normal object appearance distribution $\mathcal{N}(0, 1)$. The attributes can be specified by the user or sampled from prior class distributions of  object-attribute co-occurrence counts, which we also estimate from data during training. In this way, attribute can be treated as ``optional" at individual instance level; {\em i.e.}, one can specify sub-set of attributes for any sub-set of instances. 

\subsection{Attribute Encoder}

Visual Genome \cite{krishna2017visual} dataset provides attribute descriptions for some objects. For example, a {\tt car} object might have attributes {\tt red} and {\tt parked}, and a {\tt person} object might have attributes {\tt smile} and {\tt tall}. There are over 40,000 attributes in the datasets. We only keep the most common 106 attributes for simplicity.
In other words, $|\mathcal{A}| = 106$.
Each object might have more than one attribute, hence we adopt multi-hot embedding for the attribute encoder. If no attributes are specified for the object, the attribute embedding would be a vector of zeros, {\em i.e.}, $\mathbf{e}_i = \mathbf{0}$. We concatenate the multi-hot embedding with object word embedding and pass it through an MLP layer to obtain the final object-attribute embedding $M(\mathbf{w}_i \oplus \mathbf{e}_i)$, which is then concatenated with (prior sampled) latent code $\mathbf{z}_{obj_i}$ to construct the object feature map $F_i$.
The feature map $F_i$ is therefore constructed by filling $\mathbf{bbox}_i$ of the feature canvas with $M(\mathbf{w}_i \oplus \mathbf{e}_i) \oplus \mathbf{z}_{obj_i}$. 

\vspace{0.07in}
\noindent
{\bf Attribute Disentanglement:}
For two novel image generation paths, mainly ($L$, $Z_{obj}$) and ($L^{shift}$, $Z_{obj}$) we further entice the model to  use attribute embedding $\mathbf{e}_i$ to determine appearance of corresponding objects.
% , we want the model to use attribute embedding $e_i$ to determine the attributes of an output object. 
To explicitly disentangle attribute information from $\mathbf{z}_{obj_i}$ and $\mathbf{w}_i$ during training, we randomly choose a subset of training objects and replace their GT attributes with new attributes sampled from the attributes frequency distribution for the object class. By doing this, we force the generator to use the attributes code $\mathbf{e}_i$ to generate objects with corresponding attributes, instead of encoding attribute information into $\mathbf{z}_{obj_i}$ and/or $\mathbf{w}_i$.

% \subsection{Layout Reconfiguration}

%  In addition to image reconstruction path and image generation path, layout reconfiguration path is introduced to increase the spatial equivariance of the generator. Similar to image generation path, an object latent code $\mathbf{z}_{obj_i}$ is sampled from a normal prior distribution $\mathcal{N}(0, 1)$, and is concatenated to the object attribute embedding $M(\mathbf{w}_i \oplus \mathbf{e}_i)$. When composing the feature map $F_i^{shift}$, however, the input bounding boxes are randomly shifted. Hence, each $F_i^{shift}$ is composed based on the a new $L^{shift}$. Then, the set of $F_i^{shift}$ feature maps are downsampled and passed to a cLSTM network to form the fused map $H^{shift}$, which is then decoded back to an image $I^{shift}$. The same image discriminator is applied to the generated image $I^{shift}$, and the object discriminator, the object classifier and the attribute classifier are applied to each generated object $O^{shift}$ cropped based on the shifted bounding boxes $\mathbf{bbox}^{shift}_i$.

\begin{figure}[!t]
    \begin{center}
        \includegraphics[width=1\linewidth]{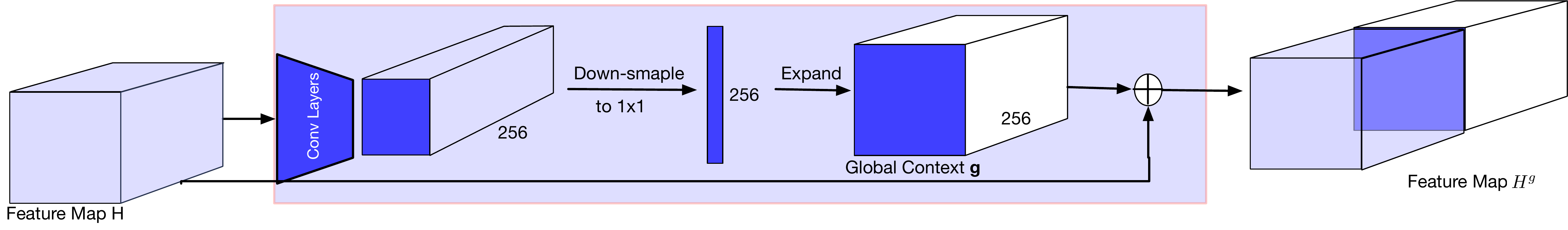}
    \end{center}
    \vspace{-0.15in}
    \caption{{\bf Global Context Encoder.} The aggregated feature map $H$ is fed into a set of conv layers; then it is down-sampled, spatially expanded and concatenated to $H$ itself to form context-aware $H^g$ feature map that can then be decoded into an image. }
    \label{fig:context_enc}
    \vspace{-0.15in}
\end{figure}

\subsection{Global Context Encoder}

At the last stage of the generation process, the fused feature map $H$ is decoded back to the output image by a stack of deconvolution layers. However, the generated image obtained from simply decoding $H$ often contains objects that are not coherent within a scene. For example, it is observed that some generated objects and the background appear incoherent and exhibit sharp transitions (image patch pasting effect). Hence, it is desirable to explicitly incorporate global context information $\mathbf{g}$ in each receptive field of the feature map $H$, so that, locally, object generation is more informed. Since $H$ contains the information for all objects, it itself is a natural choice for encoding the global context $\mathbf{g}$ (Figure \ref{fig:context_enc}). To encode $\mathbf{g}$, we feed the 8x8 feature map $H$ through two convolution layers to downsample it to a 2x2 feature map, which is average pooled to a global context vector. We then expand the vector to the size of the fused feature map $H$. 
This concatenation, $H^g = (\mathbf{g} \oplus H) $, is the final feature map used to decode the image. 
% The concatenation of $\mathbf{g}$ and $H$ is the final context-aware feature map $H^g = (\mathbf{g} \oplus H) $, which is used to decode the image. 

\subsection{SPADE normalization}

Spatially-adaptive (SPADE) normalization~\cite{park2019semantic} is an improved normalization technique that prevents semantic information from being washed away by conventional normalization layers. In SPADE normalization, the learnable transformation ({\em i.e.}, scale and shift) is learned directly from the semantic layouts. In our model, the feature map $H$ resembles the semantic layout because $H$ encodes both spatial and semantic information. Hence, we add SPADE normalization layers between the deconvolution layers in our image decoder where $H$ is used as the semantic layout. As we show later, in the ablation study (Table \ref{tb:ablation}), the object accuracy of generated results improves when we adopt SPADE. 

% \subsection{Discriminator}

% The structure of our discriminator D follows the discriminator proposed in layout2im \cite{zhao2019image}, but adds an additional term for attributes ($\mathbf{4}$).

% \begin{itemize}
% \vspace{-8pt}
%     \item[($\mathbf{1}$)] Image discriminator classifies the input image $I$ as real and the generated image $I^{rec}$, $I^{rand}$, $I^{shift}$ as fake.
% \vspace{-8pt}
%     \item[($\mathbf{2}$)] Object discriminator classifies the cropped objects $O$ from $I$ as real, and $O^{rec}$, $O^{rand}$ and $O^{shift}$ from $I^{rec}$, $I^{rand}$, $I^{shift}$, respectively, as fake.
% \vspace{-8pt}
%     \item[($\mathbf{3}$)] Auxiliary object classifier $cls^{obj}$ predicts the category of cropped objects and is used to train the generator to synthesize correct objects. It is trained on real objects $O$ and their labels $\ell$. 
% \vspace{-8pt}
%     \item[($\mathbf{4}$)] Auxiliary attribute classifier $cls^{att}$ predicts the attribute of cropped objects and is used to train the generator to synthesize objects with correct attributes. It is trained on real objects $O$ and their attributes $\mathcal{A}$. 
% \end{itemize}

\subsection{Loss Function}
The structure of our discriminator $D$ follows the discriminator proposed in layout2im \cite{zhao2019image} (see Appendices A.2 for details), but adds an additional term for attribute classifier $cls^{att}$, which predicts the attribute of cropped objects and is used to train the generator to synthesize objects with correct attributes. It is trained on real objects $O$ and their attributes $\mathcal{A}$.

Our GAN model utilizes two 
% The model follows the Generative Adversarial Networks framework \cite{goodfellow2014generative}. 
%  There are two 
adversarial losses: Image Adversarial Loss $\mathcal{L}_{adv}^{img}$ and Object Adversarial Loss $\mathcal{L}_{adv}^{obj}$. Five more losses, including KL Loss $\mathcal{L}_{\mathrm{KL}}$, Image Reconstruction Loss $\mathcal{L}_1^{img}$, Object Latent code Reconstruction $\mathcal{L}_1^{latent}$, Auxiliar Object Classification Loss $\mathcal{L}_{AC}^{obj}$  and Auxiliar Attribute Classification Loss $\mathcal{L}_{AC}^{att}$, are added to facilitate the generation of realistic images. Due to lack of space we provide details of the loss terms in Appendices Material (Section A.3). 
%
% See Supplemental 1.3 for explanations of loss terms in detail.
%
%\noindent
As the result, the generator $G$ minimizes:
\vspace{-0.05in}
\begin{equation}
\mathcal{L}_{G}= \lambda_{1} \mathcal{L}_{\mathrm{adv}}^{\mathrm{img}}+ \lambda_{2} \mathcal{L}_{\mathrm{adv}}^{\mathrm{obj}}+\lambda_{3} \mathcal{L}_{\mathrm{AC}}^{\mathrm{obj}} + \lambda_{4} \mathcal{L}_{\mathrm{AC}}^{\mathrm{att}}
+ \lambda_{5} \mathcal{L}_{\mathrm{KL}}+\lambda_{6} \mathcal{L}_{1}^{\mathrm{img}}+\lambda_{7} \mathcal{L}_{1}^{\mathrm{latent }} 
\vspace{-0.05in}
\end{equation}

\noindent
and the discriminator $D$ minimizes:
\vspace{-0.05in}
\begin{equation} \mathcal{L}_{D}= -\lambda_{1} \mathcal{L}_{\mathrm{adv}}^{\mathrm{img}} -\lambda_{2} \mathcal{L}_{\mathrm{adv}}^{\mathrm{obj}}+\lambda_{3} \mathcal{L}_{\mathrm{AC}}^{\mathrm{obj}} + \lambda_{4} \mathcal{L}_{\mathrm{AC}}^{\mathrm{att}} 
\vspace{-0.05in}
\end{equation}
where $\lambda_i$ are weights for different loss terms.

\section{Experiments}

\noindent
{\bf Datasets:}
We evaluate our proposed model on Visual Genome \cite{krishna2017visual} datasets. We preprocess and split the dataset following the settings of  \cite{johnson2018image, sun2019image, zhao2019image}. In total, we have 62,565 training, 5,506 validation and 5,088 testing images. Each image contains 3 to 30 objects from 178 categories, and each object has 0 to 5 attributes from 106 attribute set. We are unable to train on COCO \cite{lin2014microsoft} because it does not provide attribute annotations.

\begin{table*}
\small
\setlength{\tabcolsep}{2pt}
\begin{center}
\begin{tabular}{|l|c|c|c|c|c|c|}
\hline
\multicolumn{1}{|c|}{\textbf{Method}} & \textbf{FID} & \textbf{Inception} & \textbf{Obj Acc} & \textbf{Diversity} & \textbf{Attribute Score} $\uparrow$ & \textbf{Consistency} $\uparrow$ \\
 ($64 \times 64$) & $\downarrow$ & $\uparrow$ & $\uparrow$ & %$\downarrow$
 & Recall \quad Precision & bg \quad fg  \\
\hline\hline
Real Images & - & 13.9 $\pm$ 0.5  & 49.13  & - & 0.30 \qquad 0.31 & -\\

sg2im \cite{johnson2018image} & 74.61 & 6.3 $\pm$ 0.2  & 40.29  & 0.15 $\pm$ 0.12  & 0.07 \qquad 0.15 & 0.87 \qquad 0.84 \\
 Itr. SG \cite{ashual2019specifying} & 65.3 & 5.6 $\pm$ 0.5   & 28.23  & 0.18 $\pm$ 0.12  & 0.04 \qquad 0.09 & 0.82 \qquad 0.81\\
% sg2im with attribute change& 4.26 \pm 0.025  & 17.3  & 0.15 \pm 0.12  & 0.07 \qquad 0.15\\
layout2im \cite{zhao2019image} & 40.07 & 8.1 $\pm$ 0.1  & 48.09  & 0.17 $\pm$ 0.09  & 0.09 \qquad 0.22 & 0.87  \qquad  0.85 \\
LostGAN \cite{sun2019image} & 34.75 & \textbf{8.7} $\pm$ \textbf{0.4}  & 27.50  & 0.34 $\pm$ 0.10  & 0.17 \qquad 0.06 & 0.63 \qquad 0.61\\

\hline
\hline
Ours & \textbf{33.09} & 8.1 $\pm$ 0.2 & \textbf{48.82} & \begin{tabular}{l}
    0.10 $\pm$ 0.02 \\
    0.20 $\pm$ 0.01
  \end{tabular} & \textbf{0.26} \qquad \textbf{0.30} & \textbf{0.90} \qquad \textbf{0.89} \\
\hline
\end{tabular}
\vspace{-0.10in}
\end{center}
    \caption{\textbf{Performance of 64 $\times$ 64 image generation} on Visual Genome \cite{krishna2017visual} dataset)  For Diversity Score of our model, we have two versions of attribute use: GT attribute specification (top), and sampled attributes from prior class distributions of object-attribute co-occurrence counts (bottom). For Attribute Score, we predict the attributes of generated objects and calculate recall and precision against GT attributes. We trained Interactive SG without the GT object mask. $\uparrow$: higher is better; $\downarrow$: lower is better; bg: background, fg: foreground.}
    \label{tb:64x64_results}
    \vspace{-0.1in}
\end{table*}

\vspace{0.06in}
\noindent
{\bf Experimental setup:}
Our experiments use the pre-trained VGG-net \cite{simonyan2014very} as the base model to compute the inception score for generated image. For object classification loss and the attribute classification loss,  our experiments adopted the ResNet-50 model \cite{he2016deep} and replace its last $fc$ layer with the corresponding dimensions. Both object and attribute classifier are trained using the objects cropped from real training images. For attribute accuracy, we estimate the attributes of generated objects using a separately trained attribute classifier which consists of five residual blocks, and compute the recall and precision against the GT attribute annotations. Lastly, we generate two sets of test images and use LPIPS metric \cite{zhang2018unreasonable} to compute the diversity score. More specifically, we take the activation of conv1 to conv5 from AlexNet \cite{krizhevsky2012imagenet},  and normalize them in the  channel  dimension  and  take  the  L2  distance.   We  then  average across  spatial  dimension  and  across  all  layers  to  get  the  LPIPS  metric. $1$-LPIPS metric is also used for consistency score, where we compute the foreground diversity between each object before and after it is shifted, and compute the background diversity for the rest of the image which did not undergo the shift. Higher consistency for both is better.  
% The consistent or spatially equivariant generation model should have high consistency on both. 

\vspace{0.06in}
\noindent
{\bf Baselines:}
We compare our model with Sg2Im \cite{johnson2018image}, Interactive Scene Graph \cite{ashual2019specifying}, Layout2im \cite{zhao2019image} and LostGAN \cite{sun2019image}.

\subsection{Quantitative Results}

Table \ref{tb:64x64_results} and \ref{tb:128x128_results} shows the image generation results when trained using different models. For 64 $\times$ 64 images, our attribute-guided image generation from layout outperforms all other models in terms of object accuracy, attribute score and consistency score. Notably, our attribute classification score (recall and precision) outperform other models with a substantial margin, demonstrating our model's capability to control the attributes of generated objects. For consistency in layout reconfiguration, our consistency score is the highest for both background and foreground in the generated images, reflecting the effectiveness of the layout reconfiguration path. Note, as expected, specifying attributes limits the diversity of output objects ($0.10 \pm 0.02$). However, sampled from prior class distributions of object-attribute co-occurrence counts leads to much higher diversity of generated images ($0.20 \pm 0.01$).
% , but we show our model's capability to generate diverse objects if attributes are sampled from prior class distributions of object-attribute co-occurrence counts. 

% Note that our model is expected to have low diversity score due to the specification of attributes which limits the output diversity, and that the lower the diversity score for consistency the better. 

We also conduct experiments at 128 $\times$ 128 resolution and compare with LostGAN \cite{sun2019image}. %Similarly, 
Our model obtains better results on the object accuracy, attribute score and consistency score.

\subsection{Qualitative Results}

\begin{table*}[t]
\small
\setlength{\tabcolsep}{2pt}
\begin{center}
\begin{tabular}{|l|c|c|c|c|c|c|}
\hline
\multicolumn{1}{|c|}{\textbf{Method}} & \textbf{FID} & \textbf{Inception} & \textbf{Obj Acc} & \textbf{Diversity} & \textbf{Attribute Score} $\uparrow$ & \textbf{Consistency} $\uparrow$ \\ 
 (128 $\times$ 128) & $\downarrow$ & $\uparrow$ & $\uparrow$ & % $\downarrow$ 
 & Recall \quad Precision & bg \quad fg \\
\hline\hline
Real Images & - & 26.15 $\pm$ 0.23  & 41.92  & - & 0.27 \qquad 0.27 & - \\
LostGAN \cite{sun2019image} & \textbf{29.36} & \textbf{11.1} $\pm$ \textbf{0.6}  & 25.89  &  0.43 $\pm$ 0.09  & \textbf{0.19} \qquad 0.04 & 0.54 \qquad 0.51\\
\hline
\hline
Ours & 39.12 & 8.5 $\pm$ 0.1 & \textbf{31.02} & 0.15 $\pm$ 0.09 & 0.10 \qquad \textbf{0.25} & \textbf{0.84} \qquad \textbf{0.80} \\
\hline
\end{tabular}
\end{center}
\vspace{-0.1in}
\caption{\textbf{Performance of 128 $\times$ 128 images} on Visual Genome \cite{krishna2017visual} dataset. We note that images generated by LostGan \cite{sun2019image} contains too many attributes signals, which explains its high recall and low precision. $\uparrow$: higher is better; $\downarrow$: lower is better.}
\label{tb:128x128_results}
\vspace{-0.15in}
\end{table*}

\begin{table*}[!t]
\small
\setlength{\tabcolsep}{2pt}
\centering
% \vspace{+3mm}
\begin{tabular}{|l|c|c|c|c|c|}
\hline
\textbf{Method}     & \textbf{Inception}   & \textbf{Accu.}  & \textbf{Diversity} & \textbf{Attribute Score} $\uparrow$ & \textbf{Consistency} $\uparrow$ \\
  (64 $\times$ 64)  & $\uparrow$ & $\uparrow$ & % $\downarrow$ 
  & Recall \quad Precision & bg \quad fg \\\hline \hline
% w/o clstm          &    &       & \\
w/o attribute specification   & 7.9 $\pm$ 0.05   & 48.01 & 0.19 $\pm$ 0.08 & 0.08 \qquad 0.13 & 0.88 \quad 0.87\\
w/o location change    & 7.8 $\pm$ 0.1   & \textbf{48.96} & 0.12 $\pm$ 0.05 & 0.25 \qquad 0.30 & 0.86 \quad 0.84 \\
w/o SPADE \cite{park2019semantic}    & 7.9 $\pm$ 0.1   & 45.05 & 0.15 $\pm$ 0.07 & 0.23 \qquad 0.29 & 0.89 \quad 0.88 \\
w/o context encoder   & 7.7 $\pm$ 0.1   & 47.96 & 0.13 $\pm$ 0.15 & 0.24 \qquad 0.30 & 0.89 \quad 0.87 \\\hline\hline
full model    & \textbf{8.0} $\pm$ \textbf{0.2}   & 48.82 & 0.10 $\pm$ 0.02 & \textbf{0.26} \qquad \textbf{0.30} & \textbf{0.90} \quad \textbf{0.89} \\\hline
\end{tabular}
\vspace{0.05in}
\caption{\textbf{Ablation study of our model} on Visual Genome \cite{krishna2017visual} dataset by removing different objectives. Inception is the inception score, Accu. is the object classification accuracy, and Diversity is the diversity score.  $\uparrow$: higher is better; $\downarrow$: lower is better.}
\label{tb:ablation}
\end{table*}

% \begin{figure}[!t]
%     \begin{center}
%         % \fbox{\rule{0pt}{4in} \rule{0.9\linewidth}{0pt}}
%         \hspace{-20pt}
%         \includegraphics[width=1\linewidth]{images/figures/add_attribute.pdf}
%     \end{center}
%     \vspace{-0.15in}
%     \caption{\textbf{Examples of 64 $\times$ 64 generated images with added attributes on different objects} on Visual Genome \cite{krishna2017visual} datasets obtained by our proposed method.}
%     \label{fig:compare_results2}
%     \vspace{-0.1in}
% \end{figure}

% \begin{figure*}[!t]
%     \begin{center}
%         % \fbox{\rule{0pt}{4in} \rule{0.9\linewidth}{0pt}}
%         \hspace{-20pt}
%         \includegraphics[width=1\linewidth]{images/figures/64results.pdf}
%     \end{center}
%     \vspace{-0.2in}
%     \caption{\textbf{Examples of 64 $\times$ 64 generated images} on Visual Genome \cite{krishna2017visual} datasets obtained by our proposed method.}
%     \label{fig:64results}
% \vspace{-0.1in}
% \end{figure*}

Figure \ref{fig:compare_results} demonstrates our model's ability to control the attributes of generated objects. For each image, we pick an object and replace its current attribute with a different one, while keeping the rest of the layout unchanged. It can be seen from Figure \ref{fig:compare_results} that our model is able to change the attributes of the object of interest, and keep the rest of %other parts of 
the image unmodified.
% For images in Figure \ref{fig:compare_results2}, we choose objects without any attribute annotations, and assign new attributes to them. As Figure \ref{fig:compare_results2} shows, the assigned attribute are reflected on the various generated objects.

Figure \ref{fig:64locations} compares the results before and after some object bounding boxes in the canvas are horizontally shifted. For each images pair, the image on the left is generated from the GT layout, and the image on the right from the reconfigured layout. Our model shows better layout reconfigurability than other methods. For example, in Figure \ref{fig:64locations}(b') the boat is moved, in (d') two human are moved, and in (e') the horse is moved. In contrast, layout2im \cite{zhao2019image} and LostGan \cite{sun2019image} either change the theme of the image (see \ref{fig:64locations}(f')), or have missing objects (see \ref{fig:64locations}(d')) for reconfigured layouts.  This is also reflected in their much lower consistency score. 

Additional examples at 128 $\times$ 128 resolution are in Appendices, Figure \textcolor{red}{6} and \textcolor{red}{7}. 
Appendices Figure \textcolor{red}{9} shows generated images obtained using different SoTA models. Our method consistently outperforms baselines in quality and consistency of generated images. LostGan \cite{sun2019image}  fails to generate plausible human faces, and layout2im \cite{zhao2019image} does not generate realistic objects such as food. 

%  Figure 2 and 3 in Supplemental show generated images using different models. As compared to other methods, our model can generate visually appealing images for various kinds of layouts. In comparison, the latest state-of-the-art LostGan \cite{sun2019image}  fails to generate plausible human faces, and layout2im \cite{zhao2019image} does not generate realistic objects such as food. 

\vspace{-0.03in}
\subsection{Ablation Study}

We demonstrate the necessity of our key components by comparing scores of several ablated models trained on Visual Genome \cite{krishna2017visual} dataset.
As shown in Table \ref{tb:ablation}, removing any components is detrimental to the model's performance. Not surprisingly, attribute specification is the key to the successful attribute classification. The absence of layout reconfiguration path decreases the inception score by $0.2$, slightly increases the classification accuracy and, more importantly, reduces the consistency for reconfigured layouts. SPADE \cite{park2019semantic} is 
% seen to be 
beneficial for object classification accuracy, and global context encoder improves inception score by $0.3$. % helps to improve the inception score by $0.3$. 

\begin{figure}[!t]
    \begin{center}
        % \fbox{\rule{0pt}{4in} \rule{0.9\linewidth}{0pt}}
        \includegraphics[width=1\linewidth]{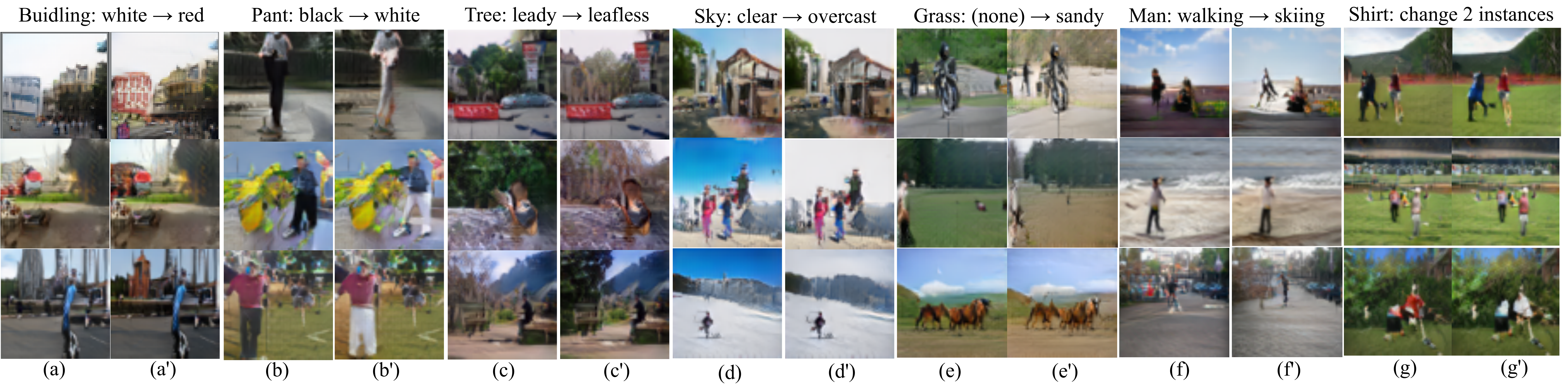}
    \end{center}
    \vspace{-0.2in}
    \caption{\textbf{Examples of 64 $\times$ 64 generated images with modified attributes} on Visual Genome \cite{krishna2017visual} datasets obtained by our proposed method.}
    \label{fig:compare_results}
    \vspace{-0.15\in}
\end{figure}

\begin{figure*}[!t]
    \begin{center}
        % \fbox{\rule{0pt}{4in} \rule{0.9\linewidth}{0pt}}
        \hspace{-20pt}
        \includegraphics[width=1\linewidth]{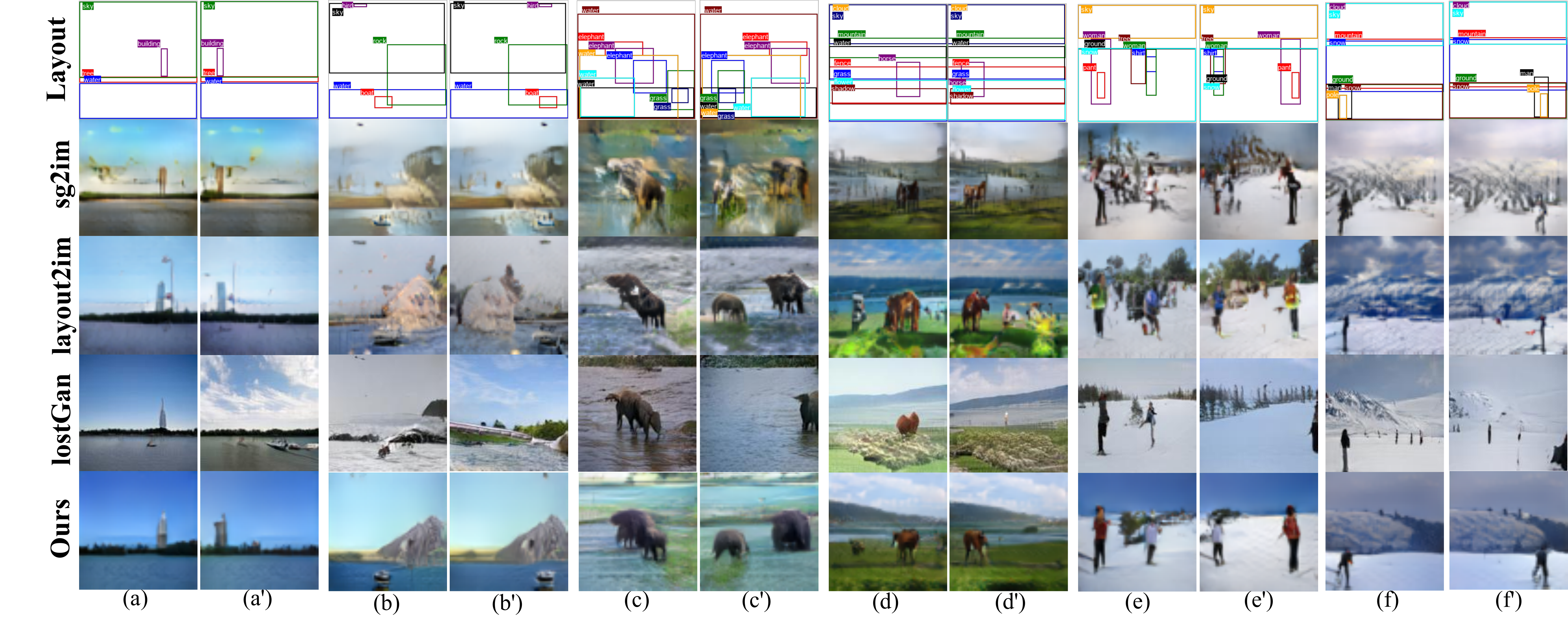}
    \end{center}
    \vspace{-0.2in}
    \caption{\textbf{Examples of 64 $\times$ 64 generated images with horizontally shifted bounding boxes} on Visual Genome \cite{krishna2017visual} datasets}
    \vspace{-0.2in}
    \label{fig:64locations}
\end{figure*}

\vspace{-0.03in}
\section{Conclusions}
\vspace{-0.03in}

This paper proposes attribute-guided image generation from layout, an effective approach to control the visual appearance of generated images in instance-level. We also showed that the model ensures visual consistence of generated images when bounding boxes in layout undergo translation. Qualitative and quantitative results on Visual Genome \cite{krishna2017visual} datasets demonstrated our model's ability to synthesize images with instance-level attribute
control and improved level of visual consistence. 

\subsection*{Acknowledgments}
\vspace{-0.05in}
This work was funded, in part, by the Vector Institute for AI, Canada CIFAR AI Chair, NSERC CRC and the NSERC DG and Discovery Accelerator Grants. Hardware support was provided by JELF CFI grant and Compute Canada under the RAC award.
\vspace{-0.1in}

\bibliographystyle{unsrt}  
\bibliography{bmvc}  %%% Remove comment to use the external .bib file (using bibtex).

\begin{appendices}
\section{Approach Details}
\vspace{-0.05in}
\subsection{Layout Reconfiguration}

 In addition to image reconstruction path and image generation path, described in the main paper, layout reconfiguration path is introduced in our model to increase the spatial equivariance of the generator. Here we describe the layout reconfiguration path a little more completely. Similar to image generation path, an object latent code $\mathbf{z}_{obj_i}$ is sampled from a normal prior distribution $\mathcal{N}(0, 1)$, and is concatenated to the object attribute embedding $M(\mathbf{w}_i \oplus \mathbf{e}_i)$. When composing the feature map $F_i^{shift}$, however, the input bounding boxes are randomly shifted. We limit ourselves to horizontal shifts in order to preserve coherence of the scene and not introduce perspective inconsistencies. Hence, each $F_i^{shift}$ is composed based on the a new $L^{shift}$. Then, the set of $F_i^{shift}$ feature maps are downsampled and passed to a cLSTM network to form the fused map $H^{shift}$, which is then decoded back to an image $I^{shift}$. The same image discriminator is applied to the generated image $I^{shift}$, and the object discriminator, the object classifier and the attribute classifier are applied to each generated object $O^{shift}$ cropped based on the shifted bounding boxes $\mathbf{bbox}^{shift}_i$.

\vspace{-0.08in}
\subsection{Discriminator}
\vspace{-0.08in}
The structure of the discriminator $D$ in our model follows the discriminator proposed in layout2im [\textcolor{green}{40}], but adds an additional term for the attributes ($\mathbf{4}$):

\begin{itemize}

    \item[($\mathbf{1}$)] Image discriminator classifies the input image $I$ as real and the generated image $I^{rec}$, $I^{rand}$, $I^{shift}$ as fake.

    \item[($\mathbf{2}$)] Object discriminator classifies the cropped objects $O$ from $I$ as real, and $O^{rec}$, $O^{rand}$ and $O^{shift}$ from $I^{rec}$, $I^{rand}$, $I^{shift}$, respectively, as fake.

    \item[($\mathbf{3}$)] Auxiliary object classifier $cls^{obj}$ predicts the category of cropped objects and is used to train the generator to synthesize correct objects. It is trained on real objects $O$ and their labels $\ell$. 

    \item[($\mathbf{4}$)] Auxiliary attribute classifier $cls^{att}$ predicts the attribute of cropped objects and is used to train the generator to synthesize objects with correct attributes. It is trained on real objects $O$ and their attributes $\mathcal{A}$. 
\end{itemize}

\subsection{Loss Function}

Our model follows the Generative Adversarial Networks framework [\textcolor{green}{4}]. Namely, one image generator and two discriminators are jointly trained in minimax game:
% with the value function V(G, D):
\vspace{-0.05in}
\begin{equation} \label{eq:gan}
\min_{G}\max_{D} %  V(D, G) =
\mathop{\mathbb{E}}_{\mathbf{x} \sim p_\mathbf{x}} [ \log D(\mathbf{x})] + \mathop{\mathbb{E}}_{\mathbf{z} \sim p_\mathbf{z}} [ \log (1 - D(G(\mathbf{z}))],
\vspace{-0.05in}
\end{equation}
where $\mathbf{x}$ is the real image sampled from the data distribution $p(\mathbf{x})$ and $\mathbf{z}$ is the latent codes that generator uses to produce fake image. Since we have two separate discriminators for images and objects, there are two adversarial losses:
\begin{itemize}
  \item \textbf{Image Adversarial Loss.}
  In each training iteration, our generator produces three images, which are: a generated image $I^{rand}$, a reconstructed image $I^{rec}$ and a shifted image $I^{shift}$. Hence, the image adversarial loss  $\mathcal{L}_{adv}^{\cdot}$ is defined as in Eq. (\ref{eq:gan}) for all three types of generated images. By averaging the loss for $I^{rec}$, $I^{shift}$, $I^{rec}$, we obtain: 
  \vspace{-0.05in}
  \begin{equation} \label{eq:L_gan_i}
  \mathcal{L}_{adv}^{img} = \frac{\mathcal{L}_{adv}^{I^{rand}} + \mathcal{L}_{adv}^{I^{rec}} + \mathcal{L}_{adv}^{I^{shift}}}{3}
  \vspace{-0.05in}
  \end{equation}
  which generator $G$ minimizes, and discriminator $D$ maximizes.

  \item \textbf{Object Adversarial Loss.}
  We crop and resize objects $O^{rand}$, $O^{rec}$ and $O^{shift}$ from $I^{rand}$, $I^{rec}$ and $I^{shift}$, respectively. By treating cropped objects as images, the object adversarial loss $\mathcal{L}_{adv}^{\cdot}$ is also defined as in Eq. (\ref{eq:gan}):
  \vspace{-0.05in}
  \begin{equation} \label{eq:L_gan_o}
  \mathcal{L}_{adv}^{obj} = \frac{\mathcal{L}_{adv}^{O^{rand}} + \mathcal{L}_{adv}^{O^{rec}} + \mathcal{L}_{adv}^{O^{shift}}}{3}
  \vspace{-0.05in}
  \end{equation}

\end{itemize}
In addition, we have another five losses to facilitate the generation of realistic images:
\begin{itemize}
  \item \textbf{KL Loss.} $\mathcal{L}_{\mathrm{KL}} = \sum_{i=1}^{o}\mathbb{E}[\mathcal{D}_{\mathrm{KL}}(Q({\mathbf{z}^r_{obj_i}} | O_{i}) \| \mathcal{N}(\mathbf{z}^r_{obj}))]$ encourages the posterior distribution $Q(\mathbf{z}^r_{obj_i} | O_i)$ for object $i$ to be close to the prior $\mathcal{N}(\mathbf{z}^r_{obj})$, for all of the $o$ objects in the given image/layout.
  
  \item \textbf{Image Reconstruction Loss.}
  $\mathcal{L}_1^{img} = \|I-I^{rec}\|_{1}$ is the L1 difference between ground-truth image $I$ and reconstructed image $I^{rec}$ produced by the generator.
  
  \item \textbf{Object Latent Code Reconstruction Loss.}
  $\mathcal{L}_1^{latent}  = \sum_{i=1}^{o} \|\mathbf{z}_{obj_i} - \mathbf{z}^{rand}_{obj_i}\|_{1} + \|\mathbf{z}_{obj_i} - \mathbf{z}^{shift}_{obj_i}\|_{1}$  penalizes the $\mathcal{L}_1$ difference between the randomly sampled $\mathbf{z}_{obj_i} \sim \mathcal{N}(\mathbf{z}_{obj})$ and the re-estimated $\mathbf{z}^{rand}_{obj_i}$ and $\mathbf{z}^{shift}_{obj_i}$ from the generated objects $O^{rand}$ and $O^{shift}$, respectively.
  
  \item \textbf{Auxiliar Object Classification Loss.}
  $\mathcal{L}_{AC}^{obj}$ is defined as the cross entropy loss from the object classifier. Cropped objects $O^{real}$ with labels from real images are used to train the object classifier, and then the generator $G$ is trained to generate realistic objects $O^{rand}$, $O^{rec}$ and $O^{shift}$ that minimize $\mathcal{L}_{AC}^{obj}$.
  
  \item \textbf{Auxiliar Attribute Classification Loss.}
  $\mathcal{L}_{AC}^{att}$ is defined as the weighted binary cross entropy loss from the attribute classifier. Similarly, real objects are used to train the classifier, and the generator $G$ is trained to generate objects $O^{rand}$, $O^{rec}$ and $O^{shift}$ with correct attribute labels that minimize $\mathcal{L}_{AC}^{att}$.

\end{itemize}

\noindent
Therefore, the generator $G$ minimizes:
% \vspace{-0.05in}
\begin{equation}
\mathcal{L}_{G}= \lambda_{1} \mathcal{L}_{\mathrm{adv}}^{\mathrm{img}}+ \lambda_{2} \mathcal{L}_{\mathrm{adv}}^{\mathrm{obj}}+\lambda_{3} \mathcal{L}_{\mathrm{AC}}^{\mathrm{obj}} + \lambda_{4} \mathcal{L}_{\mathrm{AC}}^{\mathrm{att}}
+ \lambda_{5} \mathcal{L}_{\mathrm{KL}}+\lambda_{6} \mathcal{L}_{1}^{\mathrm{img}}+\lambda_{7} \mathcal{L}_{1}^{\mathrm{latent }} 
% \vspace{-0.05in}
\end{equation}

\noindent
and the discriminator $D$ minimizes:
%\vspace{-0.05in}
\begin{equation} \mathcal{L}_{D}= -\lambda_{1} \mathcal{L}_{\mathrm{adv}}^{\mathrm{img}} -\lambda_{2} \mathcal{L}_{\mathrm{adv}}^{\mathrm{obj}}+\lambda_{3} \mathcal{L}_{\mathrm{AC}}^{\mathrm{obj}} + \lambda_{4} \mathcal{L}_{\mathrm{AC}}^{\mathrm{att}} 
%\vspace{-0.05in}
\end{equation}
where $\lambda_i$ are weights for different loss terms.

\vspace{0.1in}
\noindent
{\bf Implementation Details:}
% The object estimator, object encoder, object fuser follow the same implementation in layout2im. The object composer and image decoder are modified (see section x). 
We set image canvas size to 64 $\times$ 64 (128 $\times$ 128), and the object size to  32 $\times$ 32 (64 $\times$ 64).  The $\lambda 1$ $\sim$ $\lambda 7$ are $1.0$, $1.0$, $8.0$, $2.0$, $0.01$, $5.0$, $5.0$. The dimension of the category embedding $\mathbf{w}$ and the latent code $\mathbf{z}$ are both 64. The model is trained using Adam with a learning rate of 0.0001 and a batch size of $6$ for 300,000 iterations on 2 Geforce GTX 1080 Ti. In each training iteration, we first train the object classifier, the attribute classifier and the two discriminators, and then the generator. 
%\section{128 $\times$ 128 Images}

\section{Results}
Due to limited space in the main paper, we provide additional evaluations here. 

\subsection{Spatial Equivariance Experiments}

\begin{figure*}[!t]
    \begin{center}
        % \fbox{\rule{0pt}{4in} \rule{0.9\linewidth}{0pt}}
        \includegraphics[width=1\linewidth]{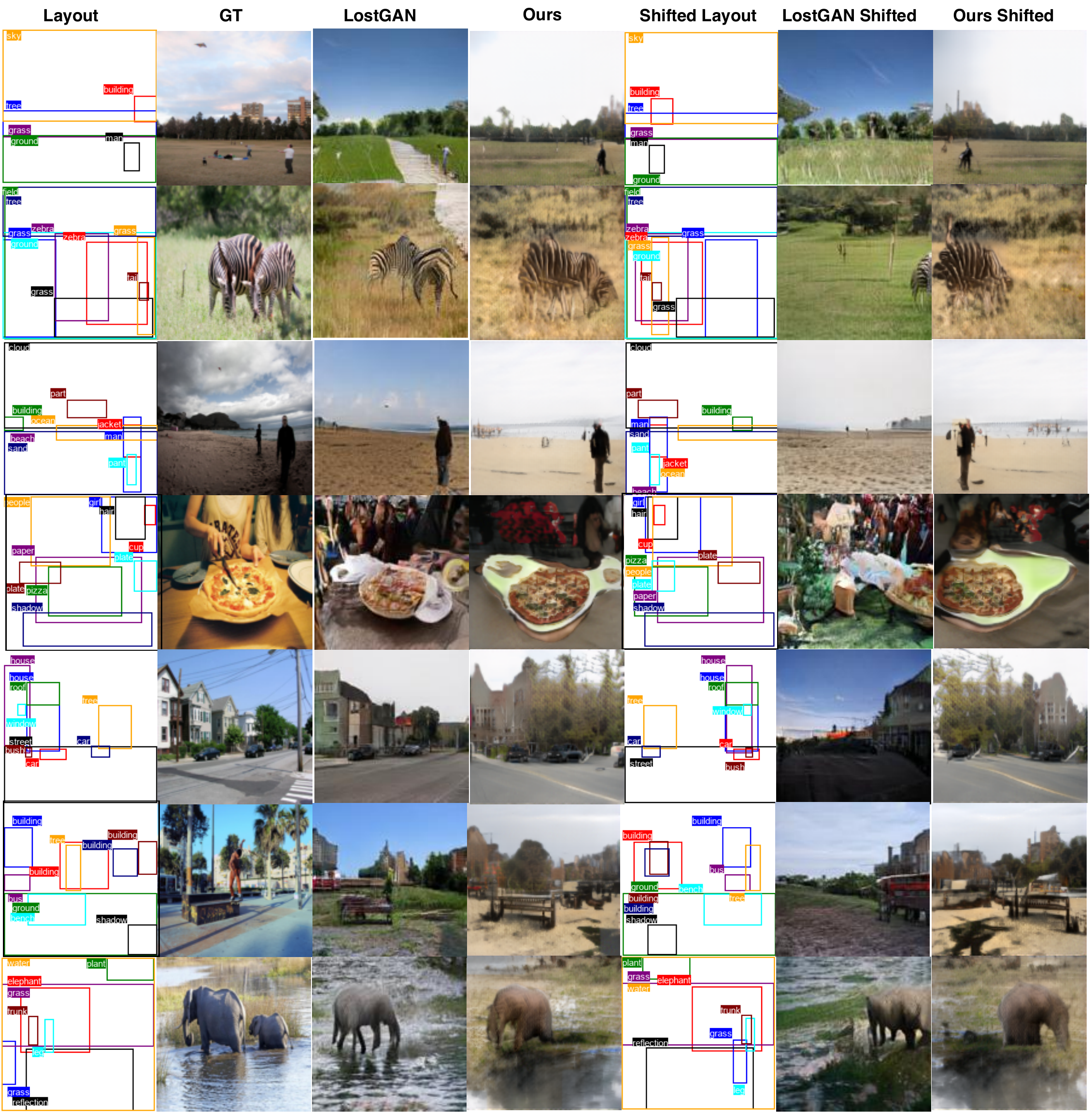}
    \end{center}
    \vspace{-0.1in}
    \caption{\textbf{Examples of 128 $\times$ 128 generated images with horizontally shifted bounding boxes} on Visual Genome datasets obtained by our proposed method.}
    \label{fig:compare_results3}
    \vspace{-0.1in}
\end{figure*}

Figure~\ref{fig:compare_results3} and~\ref{fig:compare_results} demonstrate the ability of our model to generate high quality images (at 128 $\times$ 128 resolution) and maintain consistency of objects when the boxes are shifted. We want to draw reader attention to 4-th row from the top in Figure~\ref{fig:compare_results}. Note how our model can generate images where {\tt tree} shifts from left to right based on the change in the layout (cyan), while largely maintaining the structure and appearance of the {\tt boat} unchanged. In contrast, LostGAN [\textcolor{green}{31}], when presented with the same sifted layout, generates an image that is entirely incoherent with the original: {\tt boat} is no longer discernible, sky changes color, {\em etc}. Similar behavior can also be observed in the last row, where our model is able to generate new version of the image with shifted placement of the {\tt elephants}, while maintaining the {\tt tree} line and overcast {\tt sky}. The images produced with LostGAN [\textcolor{green}{31}] are highly incoherent with visible changes in both foreground and background objects, as well as their appearances (despite fixing appearance latent vectors). Similar behavior is also readily observed in Figure~\ref{fig:compare_results3}. For example, consider new shifted placement of the {\tt person} in the third row from the top, or an almost mirror image produced by shifting {\tt trees} and the {\tt house} from right to left and vice versa in the 5-th row.  
LostGAN [\textcolor{green}{31}], while generates plausible images, is consistently failing to maintain style, appearance, structure and placement of objects when the layout is modified to simply spatially  re-arrange the same objects. 

Figure~\ref{fig:64locations} shows similar ability to maintain consistency with spatial shifts of objects in the layout at the lower, 64 $\times$ 64, resolution. Note that results of LostGAN are less blurry because, unlike all other methods in Figure~\ref{fig:64locations}, they are computed at 128 $\times$ 128 resolution (but illustrated at 64 $\times$ 64); authors of LostGAN do not provide a trained 64 $\times$ 64 model. As such, the comparison to LostGAN isn't exactly fair and is less favorable to us. Despite this, our model, is able to generate high-quality images that are consistent under spatial shifts in layout (see last row).  

%  As compared, shifting boxes in LostGAN [\textcolor{green}{31}] changes the style of generated objects significantly, and, in some extreme examples, places the objects in incorrect positions.

\begin{figure*}[!th]
    \begin{center}
        % \fbox{\rule{0pt}{4in} \rule{0.9\linewidth}{0pt}}
        \includegraphics[width=0.9\linewidth]{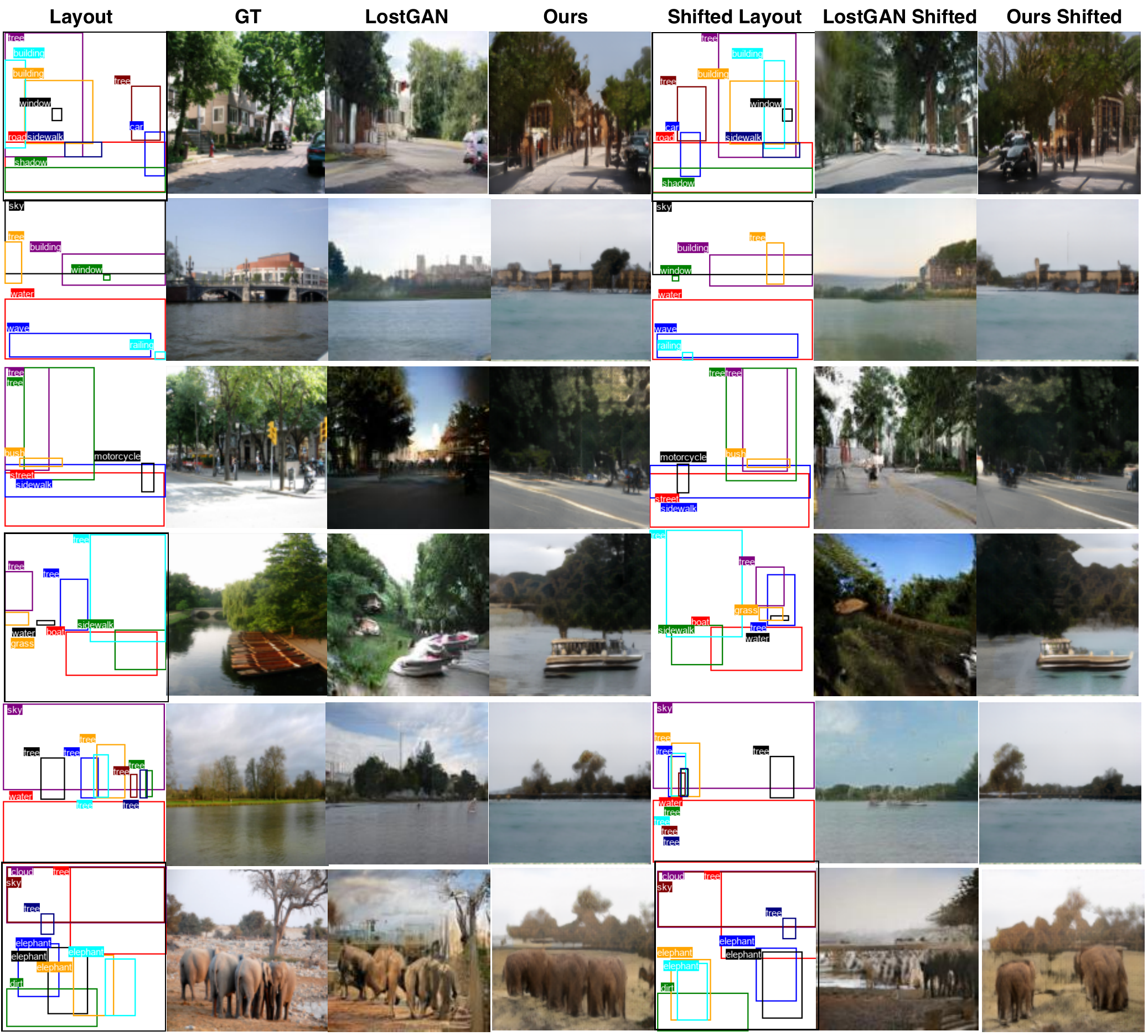}
    \end{center}
    \vspace{-0.2in}
    \caption{\textbf{Examples of 128 $\times$ 128 generated images with horizontally shifted bounding boxes} on Visual Genome datasets obtained by our proposed method.}
    \label{fig:compare_results}
    % \vspace{-0.1in}
\end{figure*}
%\newpage

\begin{figure*}[!h]
    \begin{center}
        % \fbox{\rule{0pt}{4in} \rule{0.9\linewidth}{0pt}}
        \includegraphics[width=1\linewidth]{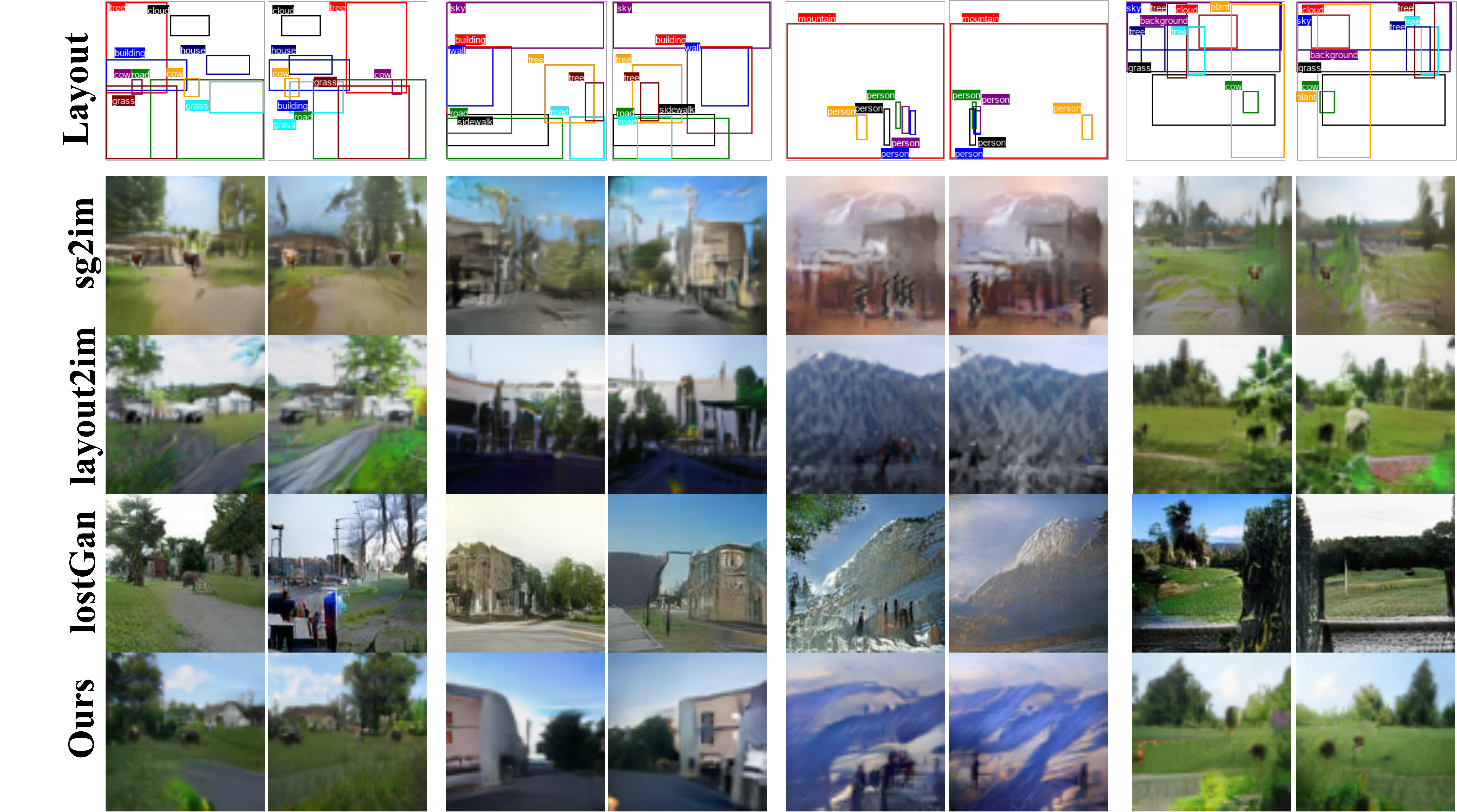}
    \end{center}
    \vspace{-0.2in}
    \caption{\textbf{Examples of 64 $\times$ 64 generated images with horizontally shifted bounding boxes} on Visual Genome datasets by our proposed method.}
    \label{fig:64locations}
\end{figure*}

\subsection{Qualitative Generation Experiments}

\begin{figure*}[!h]
    \begin{center}
        % \fbox{\rule{0pt}{4in} \rule{0.9\linewidth}{0pt}}
        \hspace{-20pt}
        \includegraphics[width=0.82\linewidth]{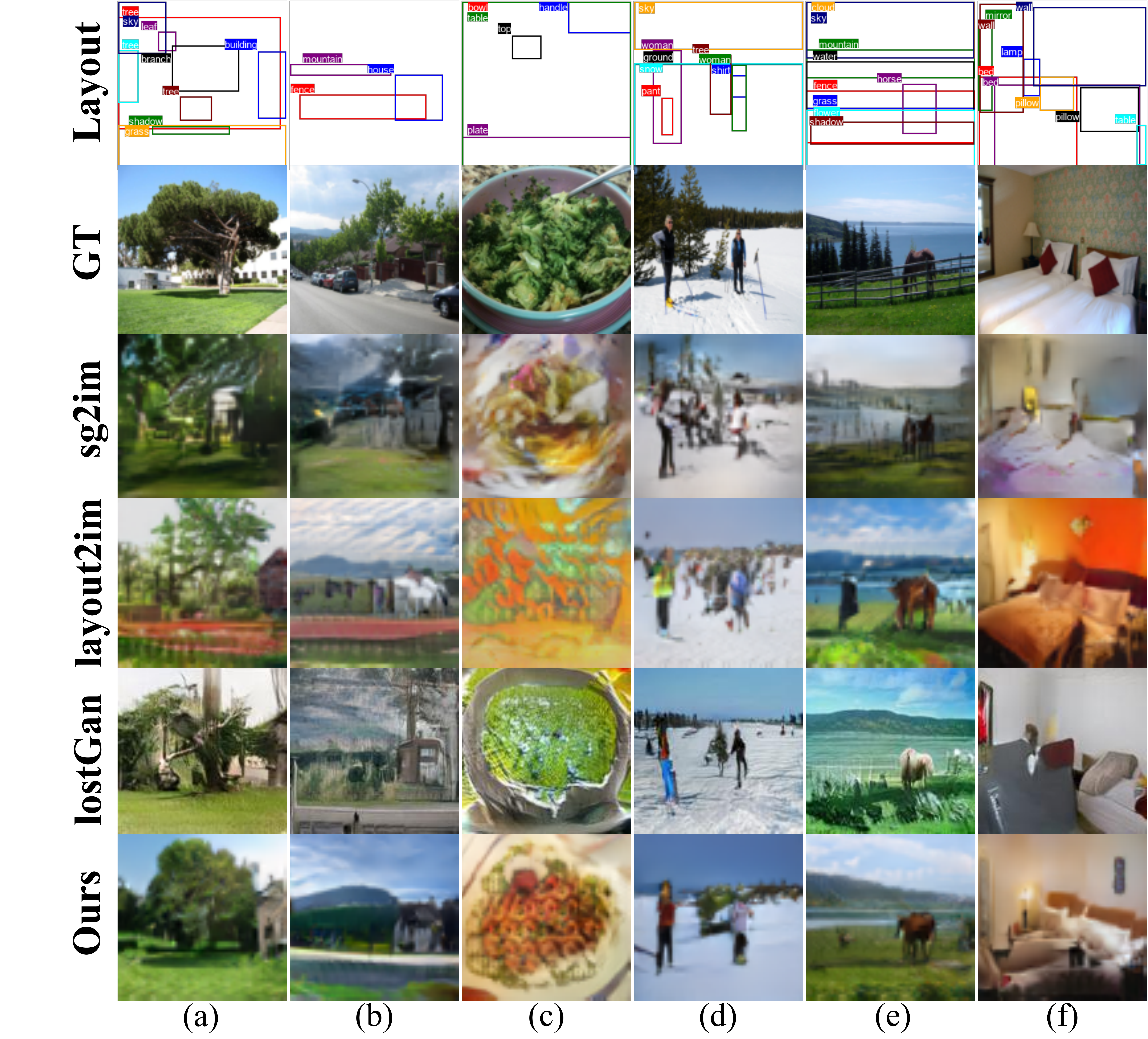}\\
         \hspace{-20pt}
        \includegraphics[width=0.82\linewidth]{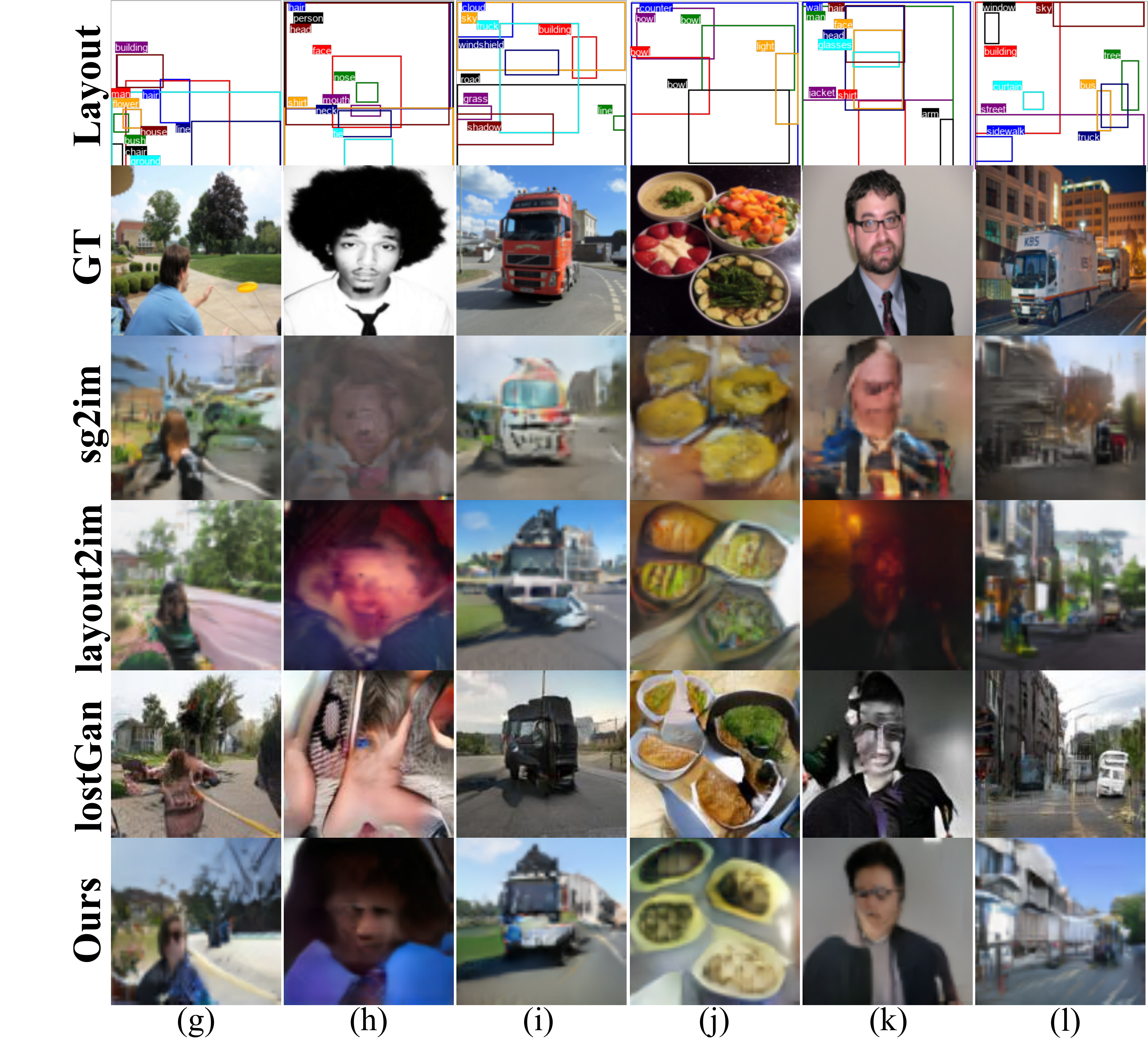}
    \end{center}
    \vspace{-0.2in}
    \caption{\textbf{Examples of 64 $\times$ 64 generated images} on Visual Genome datasets obtained by our proposed method.}
    \label{fig:64results}
\vspace{-0.1in}
\end{figure*}

Figure~\ref{fig:64results} showcases our model's ability to generate plausible images for a wide variety of layout configurations (\eg human, food, animal, furniture, house). Notably, results of sg2im [\textcolor{green}{10}] and layout2im [\textcolor{green}{40}] are of lower quality and blurry. LostGAN [\textcolor{green}{31}] does not perform well on human faces. Similar to Figure~\ref{fig:64locations}, results of LostGAN in Figure~\ref{fig:64results} are less blurry because, unlike all other methods, they are at 128 $\times$ 128 resolution; LostGAN didn't provide trained 64 $\times$ 64 model, so we use a higher resolution model instead for visualization. 

% We do provide comparison between ours and LostGAN 128 $\times$ 128 results in Figure 0 and Figure~\ref{fig:compare_results3} supplementary materials.

\subsection{Attribute Modification Experiments}

\begin{figure*}[!th]
    \begin{center}
        % \fbox{\rule{0pt}{4in} \rule{0.9\linewidth}{0pt}}
        \includegraphics[width=1\linewidth]{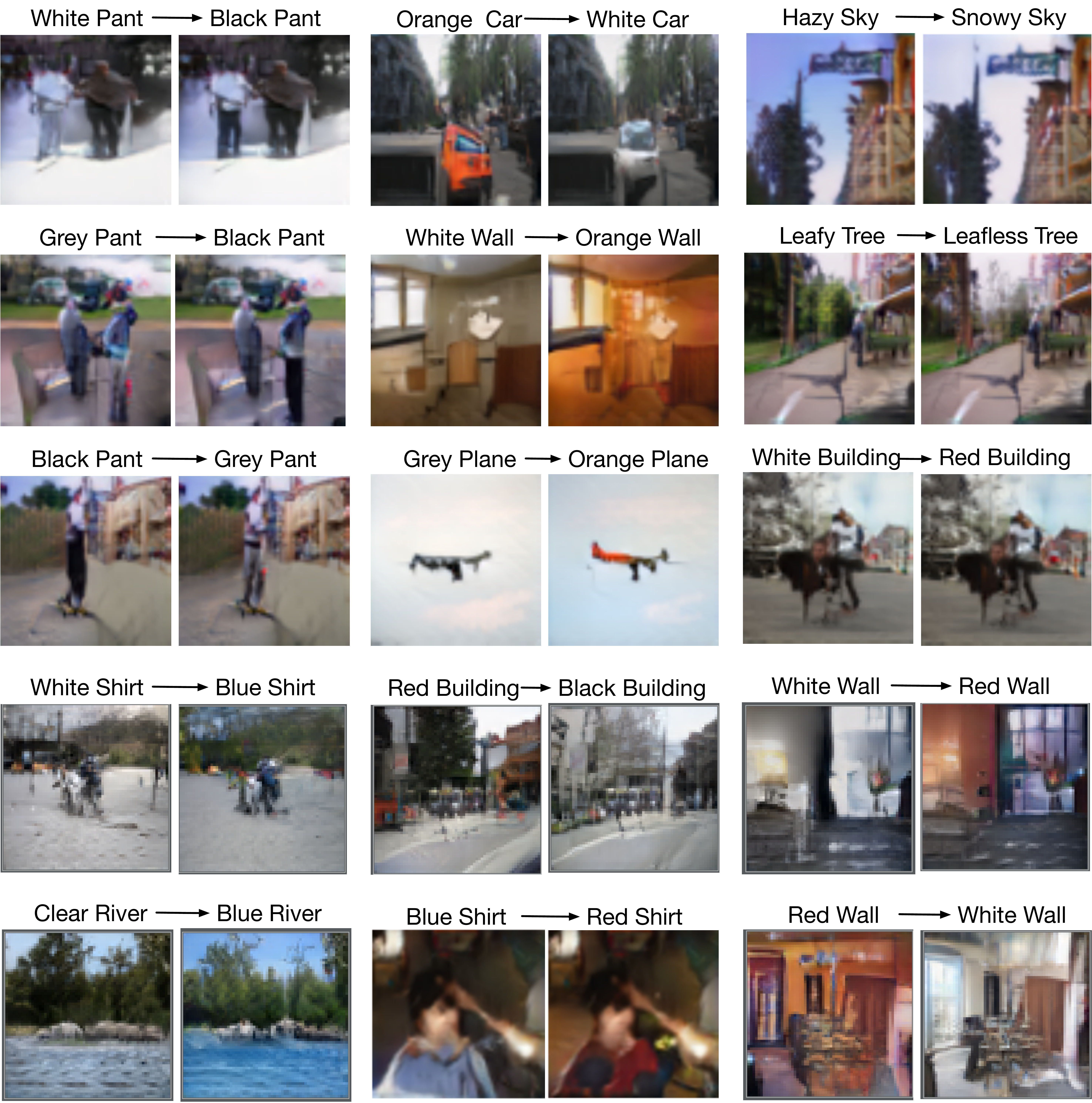}
    \end{center}
    \caption{\textbf{Examples of 64 $\times$ 64 generated images with modified attributes} on Visual Genome datasets obtained by our proposed method.}
    \label{fig:att}
    \vspace{-0.1in}
\end{figure*}

Figure~\ref{fig:att} illustrates additional examples of our model's ability to modify attributes of various objects. The change of attributes does not affect the layout or other objects in the image.

%  Figure~\ref{fig:64locations} compares the consistency of spatial shifts of object bounding boxes in the input layout. Our model is the only one that maintains the consistency and correct positions

\clearpage
\end{appendices}


\end{document}